\documentclass{article} 
\usepackage{iclr2021_conference,times}
\usepackage{caption}
\usepackage{amsmath,amsfonts,bm}

\def\eqref#1{equation~\ref{#1}}

\def\1{\bm{1}}

\DeclareMathAlphabet{\mathsfit}{\encodingdefault}{\sfdefault}{m}{sl}
\SetMathAlphabet{\mathsfit}{bold}{\encodingdefault}{\sfdefault}{bx}{n}

\newcommand{\R}{\mathbb{R}}

\usepackage{hyperref}
\usepackage{url}
\usepackage{times}
\usepackage{epsfig}
\usepackage{graphicx}
\usepackage{amsmath}
\usepackage{amssymb}
\usepackage{color}
\definecolor{Lin_color}{rgb}{0.858, 0.188, 0.478}

\definecolor{Hao_color}{rgb}{0.188, 0.478, 0.858}

\usepackage[normalem]{ulem}
\usepackage{booktabs}
\usepackage{adjustbox}
\usepackage{wrapfig}

\title{{\em Mixture of Robust Experts (MoRE)}: A Flexible \\Defense Against Multiple Perturbations}

\author{Hao Cheng$^{1}$, Kaidi Xu$^{1}$, Chenan Wang$^{1}$, Bhavya Kailkhura$^{2}$, Xue Lin$^{1}$, Ryan Goldhahn$^{2}$\\
 $^{1}$Northeastern University, $^{2}$Lawrence Livermore National Laboratory
}

\iclrfinalcopy 
\begin{document}

\maketitle

\begin{abstract}
To tackle the susceptibility of deep neural networks to adversarial examples, the adversarial training has been proposed which provides a notion of security through an inner maximization problem presenting the first-order adversaries embedded within the outer minimization of the training loss.
To generalize the adversarial robustness over different perturbation types, the adversarial training method has been augmented with the improved inner maximization presenting a union of multiple perturbations e.g., various $\ell_p$ norm-bounded perturbations. However, the improved inner maximization only enjoys limited flexibility in terms of the allowable perturbation types. In this work, through a gating mechanism, we assemble a set of expert networks, each one either adversarially trained to deal with a particular perturbation type or normally trained for boosting accuracy on clean data. The gating module assigns weights dynamically to each expert to achieve superior accuracy under various data types e.g., adversarial examples, adverse weather perturbations, and clean input. In order to deal with the obfuscated gradients issue, the training of the gating module is conducted together with fine-tuning of the last fully connected layers of expert networks through adversarial training approach. Using extensive experiments, we show that our Mixture of Robust Experts (MoRE) approach enables a flexible integration of a broad range of robust experts with superior performance. 
On CIFAR-10, when evaluating only on $\ell_p$ adversarial perturbations, we achieve up to {5.54\%, 0.40\%, and 11.79\% increase in $\ell_2 (\epsilon=1.0)$ adversarial, $\ell_\infty  (\epsilon=8/255)$ adversarial, and clean accuracy}, respectively; and when evaluating on both $\ell_p$ adversarial perturbations and adverse weather perturbations, we achieve up to {13.44\%, 9.82\%, 12.37\% and 16.92\% increase in $\ell_2 (\epsilon=1.0)$, $\ell_\infty (\epsilon=8/255)$, weather, and clean accuracy}, respectively. The code is available in \href{https://github.com/ChaduCheng/MoRE}{https://github.com/ChaduCheng/MoRE}.

\end{abstract}

\section{Introduction}
\label{sec:intro}
Deep learning has achieved exceptional performance in many application domains~\citep{he2016deep,devlin2019bert,yuan2020attribute,shi2020loss}, but its vulnerability to adversarial examples raises serious concerns \citep{szegedy2014intriguing,bulusu2020anomalous,xu2018structured,xu2019topology,xu2020adversarial,carlini2017towards}.
Recently, some success is achieved in training models  robust against a single adversary type, especially when perturbations are in an $\ell_p$-ball with a small radius surrounding the clean data point~\citep{zhang2019theoretically,shafahi2019adversarial,madry2018towards,xu2020automatic}.

To generalize the adversarial robustness over different adversary types, the adversarial training \citep{madry2018towards} method has been generalized to accommodate a union of multiple perturbations. 
Representative works in this direction include \citep{maini20classifying,maini2020adversarial,tramer2019adversarial,stutz2020confidence,dong2020adversarial, gokhale2020attribute}.
However, this line of work has limited flexibility in terms of perturbation types because sophisticated algorithms are needed to integrate multiple perturbation types into the inner maximization of adversarial training. Further, it has been observed that models trained using these methods fail to achieve robustness competitive with that of models trained on each perturbation type individually. 
Also, the robustness is gained at a significant cost -- accuracy on clean data is drastically decreased~\citep{tsipras2018robustness}.

On the other hand, model ensemble approaches have been explored for improving robustness to various attack algorithms (instead of multiple perturbation types as our paper targets).
The adaptive diversity promoting (ADP) regularizer \citep{pang2019improving} was proposed to encourage diversity, leading to robustness to different attack algorithms including FGSM \citep{Goodfellow2015explaining}, PGD \citep{madry2018towards}, C\&W \citep{carlini2017towards}, EAD \citep{chen2018ead}, etc. 
The ensemble approach has also been used for adversarial robustness to transferred attacks.
DVERGE \citep{yang2020dverge} induced diversity among sub-models for robustness to transferred attacks.
\cite{kariyappa2019improving} proposed the Gradient Alignment Loss (GAL) as the uncorrelated loss functions for an ensemble of models and thus Diversity Training for robustness under transferred attacks.
Note that these approaches do not exploit adversarial training and focus on a single perturbation type.

\textbf{Our Contributions.}

To achieve robustness under various perturbation types, in this work, we propose Mixture of Robust Experts (MoRE) method, combining adversarial training with the model ensemble approach, to achieve high adversarial accuracy as well as clean accuracy while enjoying the flexibility in perturbation types.
Specifically, we assemble a set of expert networks, including both the clean expert and robust experts targeting various perturbation types. The flexibility comes from the fact that ensemble scheme can incorporate virtually any robust expert, e.g., obtained using adversarial training against a specific perturbation type. 
\begin{wrapfigure}{r}{0.45\textwidth}
\centering
\includegraphics[scale=0.45]{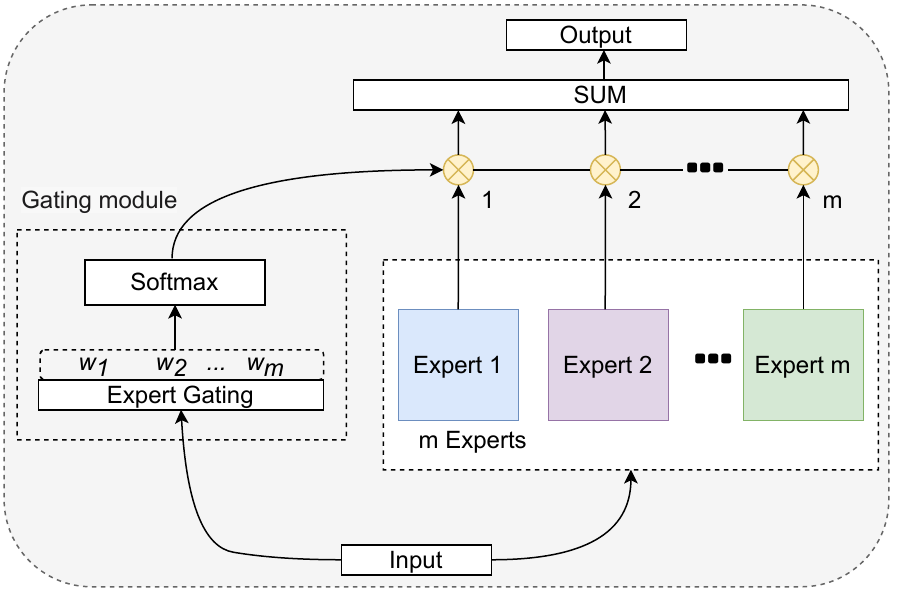}
\caption{\footnotesize{The proposed Mixture of Robust Experts (MoRE) framework with $m$ expert networks and a gating module to assign weights for the individual expert networks.}}
\label{fig_More}
\end{wrapfigure}
A gating module is designed to dynamically assign weights to expert networks during the inference.
The gating module is trained together with fine-tuning of the last fully connected layers of individual experts.
To overcome the obfuscated gradients, adversarial training is used for training the gating module and fine-tuning the experts.
Our proposed MoRE outperforms SOTA approaches on adversarial training with a union of multiple perturbation types \citep{tramer2019adversarial,maini2020adversarial}.
To the best of our knowledge, we are the first work using adversarial training within the ensemble learning paradigm to achieve adversarial robustness to different perturbation types.
Finally, the robustness to adverse weather perturbations \citep{ozdag2019susceptibility} in addition to adversarial perturbations is desired in real-world applications. Unfortunately, efforts in this direction have been virtually non-existent, and due to the flexibility of our approach can easily be supported in our framework.

\section{Proposed Mixture of Robust Experts (MoRE)}

The MoRE framework is shown in Figure \ref{fig_More} with $m$ expert networks and a gating module. 
We have three categories of \textbf{expert networks}: \emph{a clean expert}, \emph{robust experts targeting adversarial perturbations}, and \emph{robust experts targeting adverse weather perturbations}, as follows:
\begin{equation}
\small
\begin{split}\label{eq:expert_train}
    &\min_{\boldsymbol{\theta}_\text{clean}}   \mathbb{E}_{(x,y) \sim \mathcal{D}} \left[L(\boldsymbol{\theta}_\text{clean},{x},{y}) \right],\\
    &\min_{\boldsymbol{\theta}_\text{adv}}     \mathbb{E}_{({x},{y})\sim\mathcal{D}}\left[\max_{\delta\in \boldsymbol{\Delta}_p}L(\boldsymbol{\theta}_\text{adv},{x}+\delta,{y})\right],\\
    &\min_{\boldsymbol{\theta}_\text{wth}}     \mathbb{E}_{({x}_\text{wth},{y})\sim\mathcal{D}_\text{wth}}\left[L(\boldsymbol{\theta}_\text{wth},{x}_\text{wth},{y})\right],
\end{split}
\end{equation}
where $\boldsymbol{\theta}_\text{clean}$, $\boldsymbol{\theta}_\text{adv}$, and $\boldsymbol{\theta}_\text{wth}$ denote the model parameters of the clean, adversarial robust, and weather robust experts, respectively; $(x,y)\sim\mathcal{D}$ and $(x_\text{wth},y)\sim\mathcal{D}_\text{wth}$ denote the standard dataset and adverse weather dataset (which can be obtained  from standard dataset with adverse weather processing step \citep{ozdag2019susceptibility}), respectively; $\delta$ denotes the adversary perturbation within $\ell_p$-ball constraint $\boldsymbol{\Delta}_p$; and $L(\cdot)$ is the cross entropy loss.




In order to assemble clean and robust experts in an integrated robust system, we design a trainable gating module to assign weights to individual experts during inference, motivated  from Expert Gate for lifelong learning \citep{aljundi2017expert}.
Differently, with the training data, we perform training on the gating module and fine-tuning of the expert networks.
Through the joint training of the gating module and the experts, weights for individual experts are dynamically generated for each input, to achieve both high clean accuracy and high adversarial accuracy.



\noindent\textbf{Gating Module} is trained to output dynamic weights for expert networks. 
We modify the output dimension of ResNet-18~{\citep{he2016deep}} to match the number of experts in the system, as the gating module architecture.
We use  ${g}(\boldsymbol{\theta_\text{gating}}, {x})$ to denote the gating module, where $\boldsymbol{\theta}_\text{gating}$ is its parameters.

\noindent\textbf{Softmax} is added to the output of gating module to produce weights for individual experts, as:
\begin{equation}
\small
\label{eq:weight_gate}
       w = \textit{Softmax}\left({g}(\boldsymbol{\theta_\text{gating}},{x})\right).
\end{equation}




 

We use $\boldsymbol{E} \in \R^{k\times m}$ to denote the set of expert networks, where $k$ indicates the number of classes, and $\boldsymbol{E}_i$ denotes the $i$-th expert.
A total of $m$ experts are used, each one presenting either the clean expert or an adversary/weather robust expert with specific strength setting.
The output of the overall MoRE system is given by
\begin{equation}
\small
\label{eq:MoE}
{f(x) = \boldsymbol{E}\cdot w} = \sum_{i=1}^m E_i({x})\cdot w_i.
\end{equation}

\subsection{MoRE Training}

For the overall MoRE system, we train on the gating module as well as the last fully connected layers of the $m$ expert networks, with other parts of the experts fixed.
To resolve the possible obfuscated gradients issue \citep{athalye2018obfuscated} with the MoRE training, we treat the MoRE system as a whole model and train it using a modified adversarial training process, where clean training input data is feed to the whole model and PGD attacks are used to generate adversarial examples for the inner maximization. 
Different adversarial attacks are implemented alternately for the inner maximization to update the gating module and the last fully connected layers of individual experts.

\section{Experiments}\label{sec:exp}
We evaluate our MoRE system using white-box attacks: Deepfool \citep{moosavi2016deepfool} and PGD \citep{madry2018towards}, black-box attacks: RayS \citep{chen2020rays}, and adverse weather attacks \citep{ozdag2019susceptibility}.
For adversarial attacks, we consider $\ell_2$ and $\ell_\infty$ perturbations, and other $\ell_p$ perturbations can also be generalized in our method. {Please note that RayS naturally $\ell_\infty$ perturbation only. In our experiments, for a fair comparison, we add a projection step during attack to generalize Rays to both $\ell_2$ and $\ell_\infty$ constraints.}
We compare our MoRE system with SOTA approaches on adversarial training with a union of multiple perturbation types such as MAX \citep{tramer2019adversarial}, AVG \citep{tramer2019adversarial}, and MSD \citep{maini2020adversarial}.


\subsection{Experimental Setting}

We use image classification dataset CIFAR-10~\citep{Krizhevsky2009learning} in our experiments. 
ResNet18~\citep{he2016deep} is used as our base architecture for both the gating module and expert networks. 
Our MoRE system adopts a clean expert, adversarially robust experts, and weather experts.
For adversarially robust experts, we consider two $\ell_p$ perturbation types i.e., $\ell_2$ and $\ell_\infty$, and two perturbation strengths i.e., $\epsilon=$ $0.5$ and $1.0$ for $\ell_2$ experts, and $\epsilon=$ $6/255$ and $8/255$ for $\ell_\infty$ experts. 
For weather robust experts, we use two weather types, i.e., fog and snow following the adverse weather perturbation generation technique introduced in~\citep{ozdag2019susceptibility}. 
Here, $t$ and $light$ are two factors for the fog perturbations, and we set $(t, light)$ as $(0.13, 0.8)$ and  {$(0.15, 0.6)$}. Parameter $darkness$ is used to generate the snow  perturbations, and we set $darkness=2.0$ and {$darkness=2.5$}.
{When evaluating the MoRE system, we use the larger perturbation strengths} i.e., $\epsilon=1.0$  for $\ell_2$ perturbations,  $\epsilon=8/255$ for $\ell_\infty$ perturbations, $(t, light)=(0.15, 0.6)$ for fog perturbations, and $darkness=2.5$ for snow perturbations.

\subsection{Performance Evaluation}


In Table~\ref{attack_baselines}, we report the performance of our MoRE system.
We compare the performance on the clean data as well as with white-box adversaries, i.e., Deepfool \citep{moosavi2016deepfool} and PGD \citep{madry2018towards}, black-box adversary i.e., RayS \citep{chen2020rays}, and weather adversaries i.e., Fog and Snow \citep{ozdag2019susceptibility}. 
Before comparing with SOTA approaches, we report the test accuracy of a normally trained model, i.e., Column ``clean"; an $\ell_\infty$ adversarially trained model, i.e., Column ``$M_\infty$"; an $\ell_2$ adversarially trained model, i.e., Column ``$M_2$"; a fog adversarially trained model, i.e., Column ``$M_{fog}$"; and a snow adversarially trained model, i.e., Column ``$M_{snow}$".
It can be seen that these individual experts achieve high robustness on the perturbation type they are trained to be robust against. However, when presented with a different perturbation type, the robustness of these individual experts drop drastically. In contrast, MoRE achieves similar (or sometimes better) performance on all perturbation types simultaneously.

Next, we compare MoRE with MAX, AVG, and MSD baselines on $\ell_p$ adversaries only, because those SOTA methods were designed for $\ell_p$ adversaries.
We achieve up to {5.54\%, 0.40\%, 14.77\% and 11.79\% increase in $\ell_2$, $\ell_\infty$, and black-box adversarial accuracy as well as clean accuracy}, respectively.
For example, under $\ell_2$ white-box adversary Deepfool, our accuracy is 3.00\% higher than MAX; and under $\ell_2$ white-box adversary PGD, it is 5.54\% higher than AVG baseline.
Under $\ell_2$ black-box adversary RayS, our accuracy is 14.77\% higher than MSD, and under $\ell_\infty$ black-box adversary RayS, it is 5.09\% higher than MSD.

Finally, we compare MoRE with MAX and AVG on both $\ell_p$ and weather adversaries. We could not compare with MSD here, because it is not compatible to defend weather adversaries.
We achieve up to {13.44\%, 9.82\%, 18.94\%, 12.37\% and 16.92\% increase in white-box $\ell_2$ \& $\ell_\infty$, black-box, and weather adversary accuracy as well as clean accuracy}, respectively. 
For example, under fog weather adversary, our accuracy is 5.78\% higher than AVG; and under snow weather adversary, it is 12.37\% higher than AVG baseline.
Overall, our proposed MoRE achieves the best performance in almost all the evaluation cases.

\begin{table}[ht!]
\centering
 \caption{Test accuracy (\%) comparison on CIFAR-10 under different attacks: Deepfool \citep{moosavi2016deepfool} and PGD \citep{madry2018towards} as white-box adversaries; RayS \citep{chen2020rays} as black-box adversary; and Fog and Snow perturbations \citep{ozdag2019susceptibility}. For adversarial $\ell_2$ and $\ell_\infty$ perturbations, we set strength $\epsilon$ as $1.0$ and $8/255$, respectively.
 From Columns 2$\sim$6, test accuracy of clean trained, $\ell_\infty$ adversarially trained, $\ell_2$ adversarially trained, fog adversarially trained, and snow adversarially trained models are reported, respectively.
 From Columns 7$\sim$10, MoRE is compared with MAX \citep{tramer2019adversarial}, AVG \citep{tramer2019adversarial}, and MSD \citep{maini2020adversarial} on the $\ell_p$ adversaries only (without adverse weather adversaries, because these works are designed for $\ell_p$ adversaries only).
 From Columns 11$\sim$13, MoRE is only compared with MAX and AVG on both the $\ell_p$ and weather adversaries, because MSD is not compatible to defend weather adversaries.
} 
  \label{attack_baselines}
    \vspace{-0.1in}
    \adjustbox{max width=1.0\textwidth}{
\begin{tabular}{l|l|llll|llll|lll}
\toprule
\multicolumn{1}{c|}{} &

  \begin{tabular}[c]{@{}l@{}}clean \end{tabular} &
  \begin{tabular}[c]{@{}l@{}}$M_{2}$\end{tabular} &
  \begin{tabular}[c]{@{}l@{}}$M_\infty$\end{tabular} &
  \begin{tabular}[c]{@{}l@{}}$M_{fog}$\end{tabular} &
  \begin{tabular}[c]{@{}l@{}}$M_{snow}$\end{tabular} &
  \begin{tabular}[c]{@{}l@{}}{\em MAX}\end{tabular} &
  \begin{tabular}[c]{@{}l@{}}{\em AVG}\end{tabular} &
  \begin{tabular}[c]{@{}l@{}}{\em MSD}\end{tabular} &
  \begin{tabular}[c]{@{}l@{}}{\em \textbf{MoRE}}\end{tabular} &
  \begin{tabular}[c]{@{}l@{}}{\em MAX (all)}\end{tabular} &
  \begin{tabular}[c]{@{}l@{}}{\em AVG (all)}\end{tabular} &
  \begin{tabular}[c]{@{}l@{}}{\em \textbf{MoRE} (all)}\end{tabular} \\ \hline
 clean accuracy  &
   95.25&
   80.73&
   81.28&
   87.57&
   80.84&
   69.96&
   71.87&
   76.21&
   $\boldsymbol{81.75}$&
   69.88&
   66.20&
   $\boldsymbol{83.12}$\\ \hline

\begin{tabular}[c]{@{}l@{}}Deepfool $\ell_2$  ($\epsilon$ = 1.0)\end{tabular} &
   1.20&
   56.40&
   46.00&
   8.21&
   2.90&
   50.80&
   49.50&
   51.20&
   $\boldsymbol{52.50}$&
   53.50&
   44.40&
   $\boldsymbol{54.70}$\\ 
\begin{tabular}[c]{@{}l@{}}Deepfool $\ell_\infty$  ($\epsilon$ = 8/255)\end{tabular} &
   1.56&
   53.40&
   54.80&
   7.00&
   3.80&
   49.50&
   49.00&
   50.20&
   $\boldsymbol{50.60}$&
   $\boldsymbol{48.40}$&
   40.20&
   45.50\\ \hline
  \begin{tabular}[c]{@{}l@{}}PGD $\ell_2$  ($\epsilon$ = 1.0)\end{tabular} &
   0.00&
   47.97&
   35.57&
   0.15&
   0.30&
   44.13&
   43.21&
   43.92&
   $\boldsymbol{48.75}$&
   37.85&
   33.59&
   $\boldsymbol{47.03}$\\ 
\begin{tabular}[c]{@{}l@{}}PGD $\ell_\infty$  ($\epsilon$ = 8/255)\end{tabular} &
   0.00&
   39.29&
   48.01&
   0.17&
   0.28&
   42.43&
   40.30&
   $\boldsymbol{42.83}$&
   41.95&
   37.40&
   31.04&
   $\boldsymbol{40.86}$\\ \hline
\begin{tabular}[c]{@{}l@{}}RayS $\ell_2$  ($\epsilon$ = 1.0)\end{tabular} &
   1.43&
   54.40&
   53.80&
   2.00&
   6.00&
   51.67&
   54.50&
   44.73&
   $\boldsymbol{59.50}$&
   53.43&
   40.23&
   $\boldsymbol{59.17}$\\ 
\begin{tabular}[c]{@{}l@{}}RayS $\ell_\infty$  ($\epsilon$ = 8/255)\end{tabular} &
   0.71&
   40.00&
   46.50&
   0.00&
   0.00&
   43.40&
   43.21&
   41.11&
   $\boldsymbol{46.20}$&
   43.57&
   31.01&
   $\boldsymbol{46.30}$\\ \hline
   
\begin{tabular}[c]{@{}l@{}}Fog $(t, light;  0.15, 0.6)$ \end{tabular}&      
40.34&
   20.88&
   16.83&
   88.44&
   61.51&
   - &
    - &
    -&
    -&
    60.75&
    59.43&
   $\boldsymbol{65.21}$\\ 
\begin{tabular}[c]{@{}l@{}}Snow $(darkness; 2.5)$\end{tabular}&
   32.07&
   30.55&
   34.44&
   48.56&
   82.69&
    - &
    - &
    - &
    - &
    53.04&
    49.60&
   $\boldsymbol{61.97}$\\ \hline
\end{tabular}
}
\end{table}
\vspace{-0.2in}

\section{Conclusion}\label{sec:conclusion}

To achieve robustness under various perturbation types, in this work, we propose Mixture of Robust Experts (MoRE) method, combining adversarial training with the model ensemble approach, to achieve high adversarial as well as clean accuracy while enjoying the flexibility in perturbation types.
Specifically, through a gating mechanism, we assemble a set of expert networks, each one either adversarially trained to deal with a particular perturbation type or normally trained for boosting accuracy on clean data. 
In order to deal with the obfuscated gradients issue, the training of the gating module is conducted together with fine-tuning of the last fully connected layers of expert networks through adversarial training approach. Using extensive experiments, we show that our Mixture of Robust Experts (MoRE) approach enables a flexible integration of a broad range of robust experts with superior performance compared to existing  state-of-the-art approaches.

\section*{Acknowledgements} 
This work was performed under the auspices of the U.S. Department of Energy by the Lawrence Livermore National Laboratory under Contract No. DE-AC52-07NA27344, Lawrence Livermore National Security, LLC. This document was prepared as an account of the work sponsored by an agency of the United States Government. Neither the United States Government nor Lawrence Livermore National Security, LLC, nor any of their employees makes any warranty, expressed or implied, or assumes any legal liability or responsibility for the accuracy, completeness, or usefulness of any information, apparatus, product, or process disclosed, or represents that its use would not infringe privately owned rights. Reference herein to any specific commercial product, process, or service by trade name, trademark, manufacturer, or otherwise does not necessarily constitute or imply its endorsement, recommendation, or favoring by the United States Government or Lawrence Livermore National Security, LLC. The views and opinions of the authors expressed herein do not necessarily state or reflect those of the United States Government or Lawrence Livermore National Security, LLC, and shall not be used for advertising or product endorsement purposes. This work was supported by LLNL Laboratory Directed Research and Development project 20-SI-005 and released with LLNL tracking number LLNL-CONF-820522.

\bibliography{iclr2021_conference}
\bibliographystyle{iclr2021_conference}

\appendix


\end{document}